\documentclass[journal]{IEEEtran}
\usepackage{graphicx}
\usepackage{cite}
\usepackage{amsmath,amssymb,amsfonts}

\usepackage{algorithmic}
\usepackage[ruled]{algorithm2e}
\usepackage{textcomp}
\usepackage{enumerate}
\usepackage{bm}
\usepackage{color}
\usepackage{framed}
\usepackage{xcolor}
\usepackage{setspace}
\usepackage{epstopdf}
\usepackage{amsthm}
\usepackage{multicol} 
\usepackage{color}
\usepackage{subfigure}

\usepackage{tabularx}
\usepackage{booktabs}

\ifCLASSINFOpdf
\else
\fi
\begin{document}
% at, but, by, for, in, nor, of, on, or, the, to, and up, which are usually
\title{Collision-free Motion Generation Based on Stochastic Optimization and Composite Signed Distance Field Networks of Articulated Robot}

\author{Baolin Liu,
        Gedong Jiang,
        Fei Zhao$^{*}$,
        Xuesong Mei % <-this % stops a space,
\thanks{The authors are with the State Key Laboratory for Manufacturing Systems Engineering, Xi’an Jiaotong University, Shaanxi Key Laboratory of Intelligent Robots, and School of Mechanical Engineering, Xi’an Jiaotong University, Xi’an Shaanxi 710049, China (e-mail: liublhri@163.com; gdjiang@mail.xjtu.edu.cn; ztzhao@mail.xjtu.edu.cn; xsmei@mail.xjtu.edu.cn).}% <-this % stops a space
\thanks{This work was supported by the National Key R$\&$D Program of China under Grant 2022YFB3305000.}% <-this % stops a space
\thanks{}}


\markboth{-}%
{Shell \MakeLowercase{\textit{et al.}}: Bare Demo of IEEEtran.cls for IEEE Journals}



\maketitle

\begin{abstract}
Safe robot motion generation is critical for practical applications from manufacturing to homes. In this work, we proposed a stochastic optimization-based motion generation method to generate collision-free and time-optimal motion for the articulated robot represented by composite signed distance field (SDF) networks. First, we propose composite SDF networks to learn the SDF for articulated robots. The learned composite SDF networks combined with the kinematics of the robot allow for quick and accurate estimates of the minimum distance between the robot and obstacles in a batch fashion. Then, a stochastic optimization-based trajectory planning algorithm generates a spatial-optimized and collision-free trajectory offline with the learned composite SDF networks. This stochastic trajectory planner is formulated as a Bayesian Inference problem with a time-normalized Gaussian process prior and exponential likelihood function. The Gaussian process prior can enforce initial and goal position constraints in Configuration Space. Besides, it can encode the correlation of waypoints in time series. The likelihood function aims at encoding task-related cost terms, such as collision avoidance, trajectory length penalty, boundary avoidance, etc. The kernel updating strategies combined with model-predictive path integral (MPPI) is proposed to solve the maximum a posteriori inference problems. Lastly, we integrate the learned composite SDF networks into the trajectory planning algorithm and apply it to a Franka Emika Panda robot. The simulation and experiment results validate the effectiveness of the proposed method.

\end{abstract}

\begin{IEEEkeywords}
Composite SDF networks, stochastic trajectory planning, time-normalized Gaussian process prior, kernel updating strategy, collision avoidance
\end{IEEEkeywords}

% \fi
\IEEEpeerreviewmaketitle

\section{Introduction}

\IEEEPARstart{T}{he} ability to generate a safe and fast trajectory is of great significance for robot applications from industry to homes, which has attracted a lot of attention in the robotic community \cite{choset2005principles}. However, efficient and accurate collision checking for high-dimensional articulated robots is not trivial. Besides, developing optimization-based trajectory planning methods with some ability to escape local optima is also crucial for achieving this goal.

% \begin{figure}[ht]
% \centering
% \includegraphics[scale=1.2]{figs/Planning_structure.pdf}
% \caption{Illustration of signed distance Field of 3-DoFs articulated robot manipulator. The robot should have the ability to generate a collision-free trajectory from the current position to the goal position with the composite signed distance model.}
% \label{fig:1}
% \end{figure}

% , such as simple bounding volume simplification \cite{corrales2011safe}, Danger field representation \cite{5649124}, and distance field representation \cite{koptev2022neural}, etc.

Collision checking is a prerequisite for robots to generate collision-free trajectories, which aims to ensure that the robot does not collide with the environment or human beings, and several methods have been proposed for collision checking. One typical and effective method is the bounding volume hierarchy (BVH), which uses a set of discrete and simplified geometry primitives, e.g., spheres \cite{zucker2013chomp}, boxes \cite{schmidt2021real}, and capsules \cite{8062637}, to approximate the geometry structure of the robot and the environment. Those methods divide the robot and environment into a hierarchical tree structure, which enables an efficient estimate of the distance between the simplified robot and obstacles \cite{corrales2011safe}. The estimated distances serve as safety constraints for collision-free motion planning and control. However, those approaches result in a trade-off between fidelity representation of the robot geometry and collision-checking efficacy, especially for irregularly shaped articulated robots. Using many small-sized geometry primitives to approximate the geometry of the articulated robot finely will increase the computational cost in collision checking. While strong assumptions such as over-conservative robot representations, can ensure safety while reducing computational cost but also reducing the remaining free working space of the robot \cite{8613928}. A continuous but low-fidelity Danger Field is designed analytically in \cite{5649124} \cite{2017A}, considering the joint configuration and velocity of the robot, which acts as a collision avoidance constraint in a collaborative environment.

In addition to those analysis methods, machine-learning-based techniques have also been applied for collision checking. The incremental support vector machines are used to represent the collision boundary for collision pairs in configuration space while each pair of objects need a corresponding model \cite{pan2015efficient}. The kernel perceptron-inspired learning method is proposed in \cite{das2020learning} which needs active learning and online sampling to adapt to the new environment. The gradient of this extended model is applied to robot motion planning in \cite{9747928}. The signed-distance field is an intuitive effective shape representation method, which is widely used for collision checking.
The deep feed-forward networks \cite{hornik1989multilayer} in theory can learn the fully continuous Signed Distance Field (SDF) with arbitrary precision. However, the accuracy of the approximation in practice depends on the limited number of point samples guiding the decision boundary and the limited capacity of the network due to limited computing power \cite{park2019deepsdf}. The neural networks are used to build a mapping from joint space to minimum distance for articulated robots in \cite{koptev2022neural}. A trade-off exists between computational accuracy and efficiency. Most of the learning-based methods are trained in configuration space and the training data are sampled in configuration space. As the robot with a high degree of freedom, the training samples will become large. To this end, this work proposes composite SDF networks to train a simple and accurate SDF network for each link of articulated robots at the base coordinate frame. Any 3D query points, i.e., obstacles, in the working space, can be transformed into the base coordinate frame to efficiently estimate the minimum distance.

Trajectory planning is an indirect and effective method that can generate task constraints trajectories. The sampling-based \cite{elbanhawi2014sampling} and optimization-based trajectory planning algorithms are two typical trajectory planning methods. The sampling-based trajectory planning algorithm can effectively solve high-dimensional motion planning problems, which generates the desired trajectories by connecting task-constrained samples acquired in the configuration space \cite{de2007constraint} \cite{kingston2020decoupling}. These methods can guarantee computational completeness but are computationally expensive for high-DOF robots.  The optimization-based trajectory planning methods are another alternative and effective method for generating task-constrained trajectories, which iterative solve an optimal problem subjecting to equality or inequality constraints\cite{zucker2013chomp} \cite{schulman2013finding} \cite{toussaint2014newton}. Given an initial trajectory, those methods try to find the optimal solution under the task-encoded cost functions. The sequential quadratic programs are used to tackle the problem of time-optimal and collision-free motions while taking into account end-effector and object acceleration constraints imposed by the object being transported in \cite{ichnowski2020gomp} \cite{ichnowski2022gomp}. Besides, the optimization-based trajectory planning problems can be cast as a probabilistic inference with the Gaussian process prior and task-encoded likelihood, and the goal is estimating the maximum a posteriori (MAP) trajectory \cite{mukadam2018continuous}. The Stein Variational Inference is used to plan collision-free motion planning with the Gaussian process prior in \cite{lambert2021entropy} which is also a gradient-based solution method. In this case, the Gaussian process prior encodes the correlation of the time series at the velocity or the acceleration level, which can better ensure the smoothness of the trajectory \cite{barfoot2014batch}. The performance of planning algorithms is highly dependent on the parameter choosing of the prior distribution. Thus the learned Energy-Based Models (EBM) as guiding priors are used for trajectory optimization in \cite{urain2022learning}. Trajectory planning based on the Gaussian process is an effective method to generate collision-free trajectories, but the Gaussian process prior is determined by many factors, and it is not easy to design a suitable prior distribution. When the variance of the prior distribution is small, sample trajectories with small differences will be generated, which will easily cause the optimization to fall into a local optimum, while a large variance will cause the solution to diverge. For example, when the distance between the starting point and the target point changes greatly, it is necessary to readjust the troublesome prior distribution parameters. Besides, those methods are formulating the task of trajectory planning as a unified optimization problem considering the safety, smoothness, and system dynamic constraints. The results problem is complex and hard to trade-off between different tasks. Moreover, those gradient-based solving methods are easy to stick to the local optimum and are far away from the time optimal. 

We follow the framework of Gaussian process motion planning \cite{mukadam2018continuous}, but adopt the time-normalized Gaussian process as the prior distribution, and use a larger variance as the initial value to increase the diversity of trajectory samples to escape from the local optimum and explore collision-free trajectory. The trajectories are optimized using MPPI and kernel update strategies. Based on the optimized trajectory, a feasible and time-optimal trajectory is generated using an existing time-optimal planner.

\textbf{Contribution:} In this letter, we introduce a novel stochastic collision-free motion generation method for articulated robots based on composite SDF networks. First, a robot poses independent composite SDF networks are proposed for learning the SDF of the articulated robot, which can accurately estimate the minimum distance between the robot and obstacles in real time. Second, a stochastic trajectory planning algorithm is proposed to generate a fast and collision-free trajectory for a robot that has a certain ability to escape from the local optimum. The trajectory planning algorithm is formulated as a Bayesian Inference problem with a time-normalized prior and exponential likelihood function, which is solved using the kernel updating strategies and model predictive path integral (MPPI). Finally, the learned composite SDF networks are integrated into the trajectory planning algorithms which are successfully applied to generate a collision-free trajectory for the real 7-DoFs Franka Emika Panda robot.

The remainder of the letter is laid out as follows. First, we introduce the framework of the collision-free motion generation method in Section II. How to learn composite SDF networks for an articulated robot is discussed in detail in Section III. Section IV includes the stochastic trajectory planning method. Both simulations and experiments tests of the collision-free motion generation algorithm are performed and analyzed in Section V. Finally, Section VI concludes with concluding remarks and limitations of this method.

\section{An overview of the Motion Generation Method}
\begin{figure}[ht]
\centering  
\includegraphics[scale=1.25]{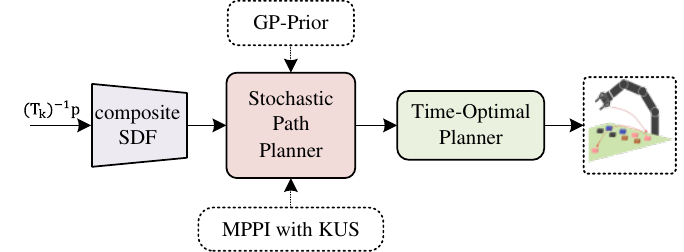}
\caption{An overview of the collision-free motion generation framework. }
\label{fig:1}
\end{figure}

The main goal of this work is to propose a collision-free motion generation method for the articulated robot which will be achieved with three cascade modules as shown in Fig. \ref{fig:1}. 

First, the composite SDF networks are learned for the articulated robot which could estimate the minimum distance from any 3D query point $\mathbf{p}$ (transform it to robot base coordinate frame based on robot inverse kinematics, $\mathbf{T}_{k}^{-1}$) to the robot in batch form. Then the estimated distance severing as obstacle avoidance constraints are streamed to a stochastic trajectory planner to generate a collision-free and spatial optimized trajectory. This stochastic trajectory planner is formulated as a Bayesian Inference problem with a time-normalized Gaussian process prior and exponential likelihood function. The Gaussian process prior is able to enforce waypoints and goal position constraints. The likelihood function is responsible for encoding containing task-related terms, such as collision avoidance, trajectory length penalty, and boundary avoidance. The model-predictive path integral (MPPI) \cite{williams2017information} combined with kernel updating strategies (KUS) are proposed to estimate the maximum a posteriori inference problems. Finally, a pre-existing time-optimal planner is used for assigning timing $t_{1:H}$ to the spatial optimized trajectory to generate spatial and temporal optimized trajectory.

\section{Composite Signed Distance Field Network for Articulated Robot}
The main goal of this section is to propose an accurate and efficient minimum distance estimation method for articulated robots based on Neural Networks.

\subsection{Composite Signed Distance File for Articulated Robot Manipulator}
Assume the pose of a $n$-DoFs robot manipulator with and $K$ links is defined by a vector of joint angles, $\boldsymbol{q} = [q_1, q_2, . . ., q_n]^{\top}\in \mathbb{R}^n$. The pose of the $k^{th}$ link $\mathbf{T}_{k}$ with respect to the base coordinate frame $\{0\}$ represented as a homogeneous transformation matrix can be easily and accurately estimated by the forward kinematics of the robot manipulator $f_{k}(\cdot)$.
\begin{equation}
    \mathbf{T}_{k} = f_{k}(\boldsymbol{q}), \quad \mathcal{T} = \{\mathbf{T}_{k}\}_{k=1}^{K}
\end{equation}
where $\mathcal{T}$ denotes a set of the homogeneous transformation matrices of all rigid links. We aim at learning a function that could accurately represent the SDF of the articulated robot manipulator using the Neural Networks. The intuitive idea is to learn a single network encoding distances directly as a map of robot configuration $\boldsymbol{q}$ and any 3D query point $\mathbf{p} \in \mathbb{R}^3$. In this way, we need to collect samples under different robot poses, and the dataset will become very large as the dimension of the robot increases. In addition, without losing accuracy, the complexity of the corresponding network needs to be increased, which also complicates training. Besides, given the joint position $\boldsymbol{q}$ of the manipulator can uniquely define the pose of each link of the robot manipulator. Thus, to simplify the learning model, instead of learning a holistic SDF network for the whole robot manipulator, we align the coordinate frame of each link to the base coordinate frame of the robot and learn an SDF network for each link.
\begin{equation}
    f_{\boldsymbol{\theta}_{k}}(\mathbf{p}) = \textbf{sdf}_{k}(\mathbf{p})
\end{equation}
where $f_{\boldsymbol{\theta}_{k}}(\mathbf{p})$ denotes the learned SDF network of link $k$ with learning parameters $\boldsymbol{\theta}_{k}$ at base coordinate frame. Given any query 3D position $\mathbf{p}$, the outputs are the approximate minimum distance to the link $k$ as well as the gradient $\hat{\mathbf{n}}=\frac{\partial  f_{\boldsymbol{\theta}_{k}}(\mathbf{p})}{\partial  \mathbf{p}}$. The SDF networks for all links are combined to form the composite SDF networks for the whole articulated robot manipulator.

\begin{figure}[ht]
\centering
\includegraphics[scale=1.2]{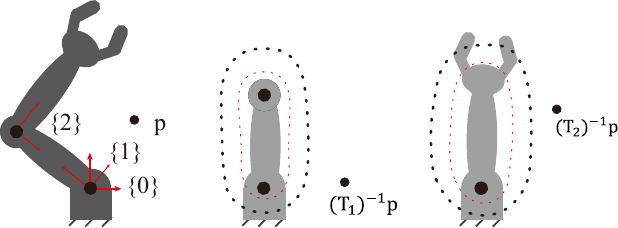}
\caption{An example of the composite signed distance field for a 2-Dof articulated robot manipulator. }
\label{fig:2}
\end{figure}

During the online inference process, firstly, the query point $\mathbf{p}$ is projected into the robot base coordinate frame $\{0\}$ by using the inverse of the homogeneous transformation matrix $(\mathbf{T}_{k})^{-1}$ as shown in \ref{fig:2}.
\begin{equation}
    \mathbf{p}_{k}^{'} = (\mathbf{T}_{k})^{-1} \mathbf{p}
\end{equation}
where $\mathbf{p}_{k}^{'}$ denotes the position with respect to the base coordinate frame of the robot. Then, the transformed point positions $\mathbf{p}_{k}^{'}$ are the input of the network.

\subsection{Dataset Synthesis and Network Designing}
In order to learn accurate and continuous SDF for the articulated robot manipulator, preparing a suitable training dataset is critical. Since the accurate 3D models of each link of the robot manipulator, such as a 3D CAD and URDF model, are easy to obtain. Thus, instead of measuring and collecting the real-world distance samples of the robot, we synthesize the datasets that capture its precise distance through the watertight 3D mesh model of each link. The training samples are collected more aggressively near the surface of the link as we are concerned with detailed distance near the robot for collision checking. First, the normal directions of the mesh vertices are estimated using Open3D library \cite{zhou2018open3d}. Those vertices whose normal vectors are inconsistent with the normal vectors of neighboring vertices are excluded. Then a set of points are sampled along the normal directions with a user-defined distance interval based on the link size. Besides, the KDtree is used to perform forward-backward tracing for rejecting the points that do not re-project on the original point \cite{liu2022regularized}. Each sample consists of the query point $\mathbf{p}$, the target distance $d$, and the normal vector $\mathbf{n}$ corresponding to the query point. using this method, the generated dataset is large enough, but under the premise of not affecting the accuracy, only a subset is randomly selected as the training set for the sake of computational efficiency. The dataset $\mathcal{D}$ for link $k$ is organized as 
\begin{equation}
    \mathcal{D}_{k} = \left\{ \mathbf{p}_{i}; d_{i}, \mathbf{n}_{i} \right\}_{i=1}^N
\end{equation}
where $\mathbf{n}_i$ is the normal vector for the $i^{th}$ sample point and $N$ is the number of samples for the $k^{th}$ link.

The fully-connected neural networks are used to learn the SDF for each robot link. Given the link number $K$, depths $D$, and the layer sizes $N$, the network complexity for the whole robot is $O(KN^{2}D)$. Since we hope to improve the calculation efficiency as much as possible without loss of the distance estimation accuracy. A deeper network will increase the computational complexity while a shallow network cannot guarantee the accuracy of distance estimation. In order to trade off accuracy and efficiency, the scale of the network is adjusted by trial and error. The final structure is composed of $5$ fully-connected layers with all hidden layers being $64$-dimensional. The ReLU function is chosen as the activation function. The loss function is defined as follows
\begin{equation}
\mathcal{L}_{\boldsymbol{\theta}_{k}} = \sum_{\mathcal{D}}\left[f_{\boldsymbol{\theta}_{k}}(\mathbf{p})-d\right]^2 + (\mathbf{t} \cdot \hat{\mathbf{n}})^2 + ( \hat{\mathbf{t}} \cdot \mathbf{n})^2
\label{eq:5}
\end{equation}
where $\hat{\mathbf{t}}$ and $\mathbf{n}$ denote the tangent of the learned SDF network and the measured normal vector at point $\mathbf{p}$, respectively. $\mathbf{t}$ and $\hat{\mathbf{n}}$ denote the measured tangent vector of the point corresponding to the link mesh vertices and the estimated normal vector at point $\mathbf{p}$. The first term aim at penalizing distance errors. The second and third terms are used to penalize the normal vector alignment errors as in \cite{liu2022regularized}.

\section{Stochastic Optimization for Robot Trajectory Planning}

The optimization-based trajectory planning problem can be formulated as a Bayesian inference problem, following the conventions established in \cite{mukadam2018continuous} and \cite{toussaint2009robot}. Mathematically, this is expressed as:

\begin{equation}
p(\boldsymbol{\tau} | \mathcal{Z}=1) \propto p(\mathcal{Z}=1 | \boldsymbol{\tau}) p(\boldsymbol{\tau})
\end{equation}
where a prior distribution on the trajectory, $p(\boldsymbol{\tau})$, encodes prior knowledge such as initial and goal constraints.  $\boldsymbol{\tau} \triangleq [\mathbf{x}_{0}, \mathbf{x}_{1}, ..., \mathbf{x}_{H}]$ is defined as a sequence of positions of the robot, where $\mathbf{x}=[x, y]^{\top} \in \mathbb{R}^{2}$ for a 2D planner mobile robot, and $\mathbf{x}=\boldsymbol{q} \in \mathbb{R}^{n}$ for an $n$-DoF robot manipulator. $H$ denotes discrete time steps. The likelihood function, $p(\mathcal{Z}=1 | \boldsymbol{\tau})$, encourages the achievement of desired objectives with  a set of optimization criteria $\mathcal{Z}$. For instance, this could involve optimizing for obstacle and joint limits avoidance. $\mathcal{Z}=1$ indicates the corresponding criterion is optimized. Thus, the objective is to find the maximum a posteriori (MAP), $p(\boldsymbol{\tau} | \mathcal{Z}=1)$, with a likelihood function that encourages the trajectories to be collision-free.

\subsection{Time-Normalized Gaussian Process Prior}
\begin{figure}[ht]
\centering
\subfigure[]{
\begin{minipage}[t]{0.47\linewidth}
\centering
\includegraphics[width=\textwidth]{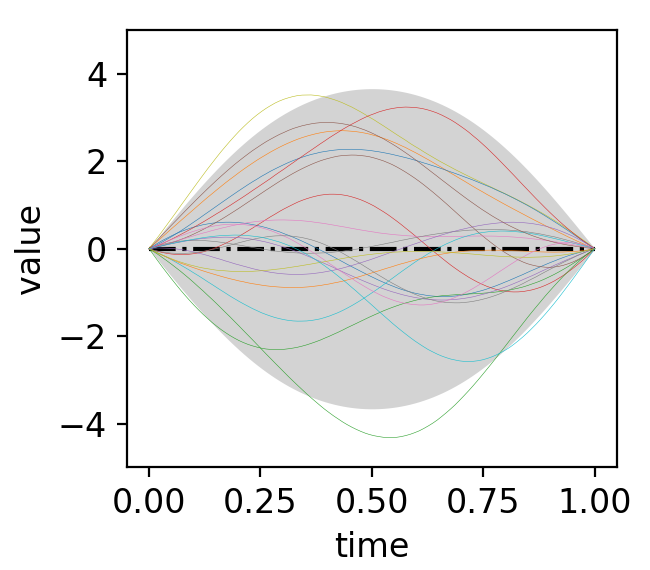}
%\caption{fig1}
\end{minipage}%
}%
\subfigure[]{
\begin{minipage}[t]{0.47\linewidth}
\centering
\includegraphics[width=\textwidth]{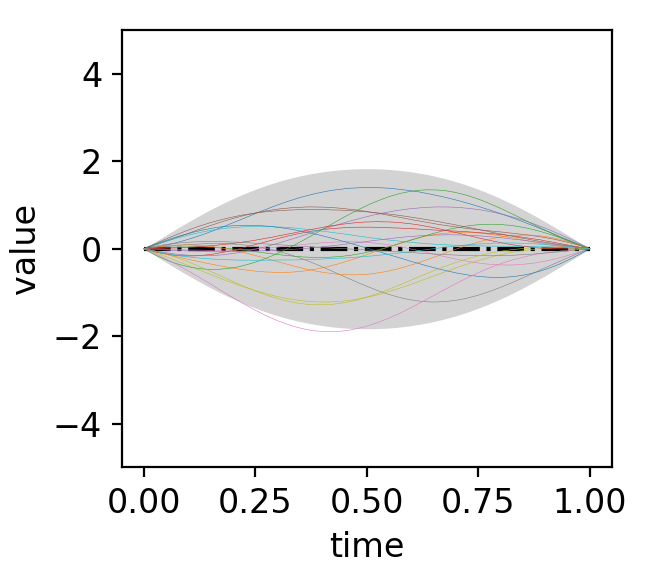}
%\caption{fig2}
\end{minipage}%
}%
\centering
\caption{An example of the time-normalized GP-prior  with different $\sigma_{f}$. $20$ solid lines are the trajectories sampled from the prior. The black dotted line represents the mean of the prior. (a) $\sigma_{f} = 4, h = 0.1$. (b) $\sigma_{f} = 1, h = 0.1$.}
\label{fig:3}
\end{figure}

Since the proper prior allows encoding some desired behaviors, to facilitate generating optimal and goal-directed motion trajectory for the robot, we wish to incorporate goal constraints and smoothness, i.e., the correlation of trajectories over time series, into the prior distribution. The main goal of this part is to generate spatial-optimized collision-free trajectories, regardless of trajectory feasibility, i.e., not considering system dynamic constraints here. Besides, instead of constructing Gaussian process priors at the velocity and acceleration level, the Gaussian process prior is designed at the position level to increase the diversity of trajectory samples for exploring collision-free trajectories. To this end, the simple Time-Normalized Gaussian process with the time $t$ being normalized within interval $[0, 1]$, is proposed to define a prior distribution over the trajectories $\boldsymbol{\tau}$. Thus, the prior distribution can is modeled as
\begin{equation}
p(\boldsymbol{\tau}) \propto \exp \left\{-\frac{1}{2}\|\boldsymbol{\tau}-\boldsymbol{\mu}\|_{\mathcal{K}}^{2}\right\}
\end{equation}
where \(\|\boldsymbol{\tau}-\boldsymbol{\mu}\|_{\mathcal{K}}^{2} \) is the Mahalanobis distance with the Gaussian kernel \(\mathcal{K}\). 
For the general goal-directed trajectory planning problem, the initial and target positions are known in advance. The kernel of the initial Gaussian process prior over trajectory $\boldsymbol{\tau}$ conditioned on the observed initial and goal positions can be estimated by the following formula
\begin{equation}
    \boldsymbol{\mathcal{K}} = \boldsymbol{\mathcal{K}}_{**} - \boldsymbol{\mathcal{K}}_{*}^{\top} \boldsymbol{\mathcal{K}}_{test}^{-1} \boldsymbol{\mathcal{K}}_{*}
\end{equation}
where the Gaussian kernel $\boldsymbol{\mathcal{K}}_{*}$, $\boldsymbol{\mathcal{K}}_{test}$ and $\boldsymbol{\mathcal{K}}_{**}$ are 
\begin{equation}
\begin{aligned}
& \boldsymbol{\mathcal{K}}_{*} = [\kappa\left(t^{j}, t^{i}\right)], \ i \in \boldsymbol{\mathcal{C}}, \ j \in \boldsymbol{\mathcal{C}}_{*} \\
& \boldsymbol{\mathcal{K}}_{test} = [\kappa\left(t^{j}, t^{i}\right)], \ i, j \in \boldsymbol{\mathcal{C}} \\
& \boldsymbol{\mathcal{K}}_{**} = [\kappa\left(t^{j}, t^{i}\right)], \ i, j \in \boldsymbol{\mathcal{C}}_{*}
\end{aligned}
\end{equation}
where $\boldsymbol{\mathcal{C}}$ denotes the set of test points indexes, i.e., the indexes of the discrete waypoints. $\boldsymbol{\mathcal{C}}_{*}$ denotes the indexes of the observed initial and goal points, i.e., $0$ and $H$. Each element of the kernel matrix is determined by
\begin{equation}
\kappa\left(t^{j}, t^{i}\right)=\sigma_{f}^{2}\exp \left\{-\frac{1}{2h^{2}} (t^{j}-t^{i})^{2}\right\}, \ t^{i}, t^{j} \in [0, 1]
\label{eq:10}
\end{equation}
where $\sigma_{f}$ and $h$ are two hyper-parameters. Therefore, the diversity of the trajectories sampled from the conditional Gaussian process prior could be easily adjusted by the parameter, $\sigma_{f}$, as shown in Fig. \ref{fig:3}.

% \begin{figure}[ht]
% \centering
% \includegraphics[scale=0.47]{figs/sigma_var.pdf}
% \caption{An example of the time-normalized GP-prior. The shaded 100 solid lines are the sample trajectories that lie within an interval with a standard deviation of $2\sigma_{max}$. The black dotted line represents the mean of the trajectory. (a) $\sigma_{f} = 4, h = 0.1$, (b) $\sigma_{f} = 1, h = 0.1$,}
% \label{fig:2}
% \end{figure}

\subsection{The Likelihood Function for Trajectory Planning}
The prior distribution presented above primarily enforces goal constraint. Moreover, we also need to define likelihood functions that encourage the achievement of other desired objectives, e.g., collision avoidance and trajectory length penalty. In this case, the Exponential Utility is chosen as the cost-likelihood to quantify each desired objective
\begin{equation}
p(\mathcal{Z} \mid \boldsymbol{\tau}) = \exp \left\{-\frac{1}{2}\sum_{i=1}^{l}\|\ell_{i}(\boldsymbol{\tau})\|_{\boldsymbol{\Sigma}_{i}}^{2}\right\}
\end{equation}
where $l$ is the number of tasks. The weight matrix  $\boldsymbol{\Sigma}_{i}$ corresponds to the $i^{th}$ task.

The cost for collision avoidance is defined as follows.
\begin{equation}
\ell_{obs}(\boldsymbol{\tau}) = \sum_{i=0}^{H}\ell_{obs}(\mathbf{x}_{i}), \ \ell_{obs}(\mathbf{x}_{i}) = \begin{cases} 1, &\text{ if} \ d(\mathbf{x}_{i}) \leq \epsilon \\ 0, &\text { otherwise } \ \end{cases}
\end{equation}
where $d(\mathbf{x}_{i})$ is the minimum distance between the robot and obstacles which could be estimated using the composite SDF networks. $\epsilon$ is the distance threshold. The cost for obstacle avoidance is estimated at a series of discrete waypoints. In order to avoid collisions with delicate objects such as sharp corners and thin-walled obstacles, the evaluation points are added between adjacent waypoints by linear interpolation when calculating the cost for obstacle avoidance.

% Given the maximum joint state limits, we need to define a cost to penalize robot states exceeding the predefined threshold. As the state exceeds the threshold, this cost-likelihood provides the robot with a high probability of acceleration pointing in the opposite direction to the joint state limits. Let define two joint state thresholds, \(\mathbf{\overline{x}}_{min}, \ \mathbf{\overline{x}}_{max} \), for the manipulator.  
% \begin{equation}
% \mathbf{\overline{x}}_{min} = \mathbf{x}_{min}+ \Delta\mathbf{x}, \quad \mathbf{\overline{x}} = \mathbf{x}_{max}-\Delta\mathbf{x},
% \end{equation}
% where \(\Delta\mathbf{x}\) is a positive increment vector of state.  \(\mathbf{x}_{max}\) and \(\mathbf{x}_{min}\) are the joint state limits. The cost term of joint state limits at each prediction time step is 
% \begin{equation}
% \ell_{joint}^{k} = \begin{cases} \|\mathbf{x}_{k}-\mathbf{\overline{x}}_{min} \|_{2}, & \text { if } \mathbf{x}_{k}<\mathbf{\overline{x}}_{min} \\ \|\mathbf{x}_{k}-\mathbf{\overline{x}}_{max} \|_{2}, & \text { if } \mathbf{x}_{k}>\mathbf{\overline{x}}_{max} \\ 0, & \text { otherwise }\end{cases}
% \end{equation}
% The cost of joint state limits during the whole prediction horizon \(\mathcal{T}\) is
% \begin{equation}
% \ell_{joint}(\boldsymbol{\tau}) = \sum_{k=0}^{T-1}\ell_{joint}^{k}
% \end{equation}

In addition to the collision avoidance tasks, a trajectory length penalty is also added to penalize the length of the trajectory. Thus, this cost is designed as
\begin{equation}
\ell_{le}(\boldsymbol{\tau}) = \sum_{i=0}^{H}\|\mathbf{x}_{i+1}-\mathbf{x}_{i}\|
\end{equation}
Other task-related cost items can be also added as required.

\subsection{Trajectory Optimization and Kernel Updating Strategy}

In this part, the goal is to estimate the maximum posterior, \(p\left(\boldsymbol{\tau} \mid \mathcal{Z}\right)\), given the above-mentioned Gaussian process prior and likelihood functions. The main idea is to continuously update this Gaussian prior so that the trajectories samples from the Gaussian prior can maximize the likelihood functions. The weighted mean trajectory is the solution result. Thus, it is necessary to update not only the mean of the Gaussian process prior but also the kernel matrix. The mean trajectory is updated using the derivation-free model predictive path integral (MPPI) \cite{williams2017information} with weight approximated by Monte Carlo sampling. Let us draw a number of $N_{s}$ trajectory samples from the Gaussian process prior, $p(\boldsymbol{\tau})$, and estimate the cost-likelihood value of each trajectory. The mean trajectory is updated as follows
\begin{equation}
\boldsymbol{\mu}_{t+1}=\boldsymbol{\mu}_{t} + \gamma \sum_{i=1}^{N_{s}} \omega_{i} (\boldsymbol{\tau}_{i}-\boldsymbol{\mu}_{t})
\label{eq:17}
\end{equation}
where the step-size $\gamma$ is used to get a smooth updating process and the relative probabilistic weight $\omega_{i}$ corresponding to the $i^{th}$ trajectory is determined according to follows
\begin{equation}
\omega_{i} = \frac{p(\mathcal{Z}\mid \boldsymbol{\tau}_{i})}{\sum_{i=0}^{N_{s}}p(\mathcal{Z} \mid \boldsymbol{\tau}_{i})}
\end{equation}

The kernel matrix, $\boldsymbol{\mathcal{K}}$, can determine the diversity of the sampled trajectories, and it can be adjusted by the parameter $\sigma_{f}$ and $h$. Since $h$ is responsible for controlling the correlation of waypoints in time series that is not the main factor affecting the exploration of collision-free trajectories. To this end, we mainly discuss the strategies of updating $\sigma_{f}$. A large $\sigma_{f}$ corresponds to a large sampling region and can results in more differentiated trajectories to find the collision-free trajectories. However, a large $\sigma_{f}$ also causes the optimization to generate conservative trajectories. To this end, we prefer to use a large $\sigma_{f}$ to initially explore collision-free trajectories, and to use a small $\sigma_{f}$ to fine-tune the trajectory after finding the collision-free trajectory.

The parameter of $\sigma_f$ is updated as follows
\begin{equation}
\sigma_{f}^{t+1} = \begin{cases} \sigma_{f}^{t}, & \text { if } \ell_{obs}(\boldsymbol{\mu}_{t}) > 0  \\ \eta \cdot \sigma_{f}^{t}, & \text { if } \ell_{obs}(\boldsymbol{\mu}_{t}) = 0 \text { and } \sigma_{f}^{t} > \sigma_{min} \\ \sigma_{min}, & \text { otherwise }\end{cases}
\end{equation}
where $\sigma_{f}^{t}$ and $\sigma_{f}^{t+1}$ denote the standard deviation of the current and next iteration step. $\eta$ is an attenuation factor, $0<\eta<1$. $\ell_{obs}(\boldsymbol{\mu}_{t})$ is the cost value for obstacle avoidance corresponding to the mean trajectory $\boldsymbol{\mu}_{t}$. $\sigma_{min}$ refers to the threshold of standard deviation.

% The number of discrete waypoints $H$ used to parameterize the path is determined by the complexity of tasks. A higher value will generate finer trajectories but also increases computational complexity.

\subsection{Time-Optimal Path Parameterization}
The above-generated trajectories are spatially optimal, while it is not feasible, i.e., without considering system dynamics. Thus it is necessary to allocate the timing for the trajectories to generate time-optimized trajectories while considering the system dynamics, velocity, and acceleration constraints. The off-the-shelf Reachability Analysis-based time-optimal path parameterization algorithm (TOPP-RA) is used to generate a time-optimal trajectory \cite{pham2018new}. The TOPP-RA algorithm uses a cubic spline curve to fit the waypoints which also ensures the smoothness of the trajectory. As a result, the output of the time-optimal planner is the final trajectory which is not only spatial optimal but also temporal optimal.

% \begin{algorithm}
% \caption{Stochastic Optimization for Robot Trajectory Planning}
% % \LinesNumbered % line index
% \KwIn{input parameters: current state $\mathbf{x}_{t}$, goal state $\mathbf{x}_{g}$}% input parameters
% \KwOut{trajectory $\boldsymbol{\tau}^{*}$}% output parameters;  %\; linefeed

% \While{Task not complete}{

% Sample trajectories  \( \{\mathbf{\tau}_{t}^{i} \}_{i=1}^{N} \sim \mathcal{GP}(\boldsymbol{\mu}, k(t,t')) \) \;

% }
% \end{algorithm}

\section{Simulation and Experiment Validation}
Both simulations and experiments have been carried out to verify the effectiveness and performance of the collision-free motion generation algorithm. First, we tested the computational efficiency and accuracy of the composite SDF network. Then, the trajectory planning algorithm is tested with a 2D mass point. Finally, the learned SDF networks are integrated into the trajectory planning algorithm with the Franka Emika Panda robot to generate collision-free trajectories. All the algorithms are implemented in Python and Pytorch which allows for batch operations and run on a laptop with AMD Core i7-9700 and Geforce RTX 3060 GPU.

\subsection{Evaluation of Composite SDF Networks}
The composite SDF Networks are trained for the Franka Emika Panda robot which only takes a few minutes before converging for each link. This is very convenient for adjusting the hyper-parameters and can be easily applied to other robots. Both the accuracy and computational time of the learned SDF networks are evaluated.

\begin{table}[ht]
\centering
\caption{Computational time of the learned SDF network.}
\begin{tabularx}{\linewidth}{l|X|X|X|X|X}
\toprule
$ N_{p} $ & 1 & 10 &100 & 1000 & 10000  \\
\midrule
$ t_c(ms) $ & 0.532 & 0.551 & 0.568 & 0.641 & 0.725  \\
\bottomrule
\end{tabularx}
\label{tab:1}
\end{table}%

The average query computation time with different numbers of 3D query points for each link is investigated in a batch fashion on an RTX 3060 GPU, and the results are listed in Table. \ref{tab:1}. $N_{p}$ is the number of test points and $ t_c $ is the computational time for each link. The computational time increases as the number of query points increases. Even if the number of query points is 10000, the calculation time is less than $1$ ms.
% As the number of query points is less than 1000, the calculation time is less than $1$ ms. The number of query points is 10000, and the calculation time is $2.563$ ms.
% which is up to 100x faster than the standard Gilbert-Johnson-Keerthi (GJK) algorithm. 
% Thus, this model is computationally efficient that meets most real-time obstacle avoidance application scenarios of the robot manipulator. 

\begin{figure}[ht]
\centering
\subfigure[]{
\begin{minipage}[t]{0.32\linewidth}
\centering
\includegraphics[width=\textwidth]{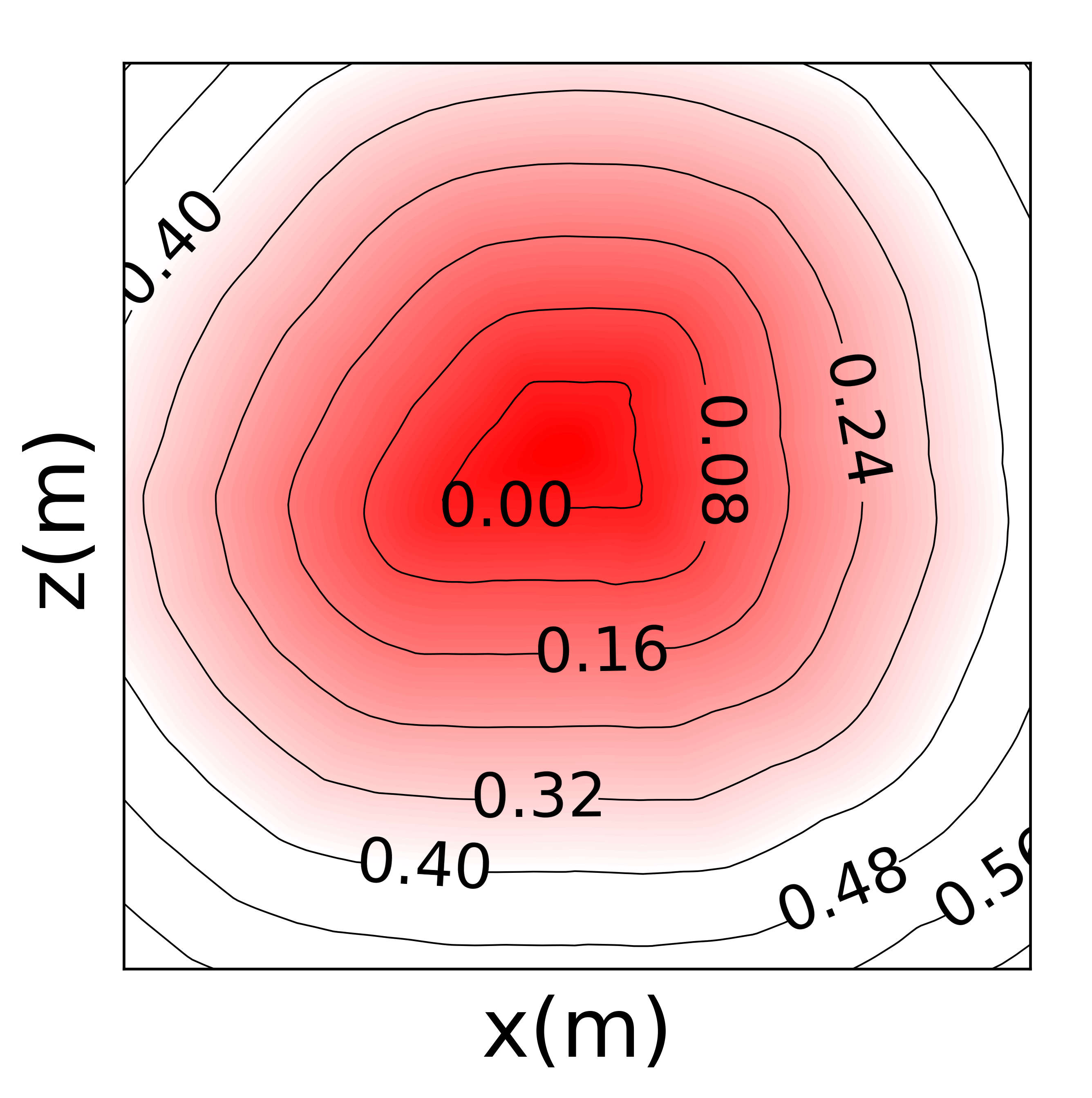}
%\caption{fig1}
\end{minipage}%
}%
\subfigure[]{
\begin{minipage}[t]{0.32\linewidth}
\centering
\includegraphics[width=\textwidth]{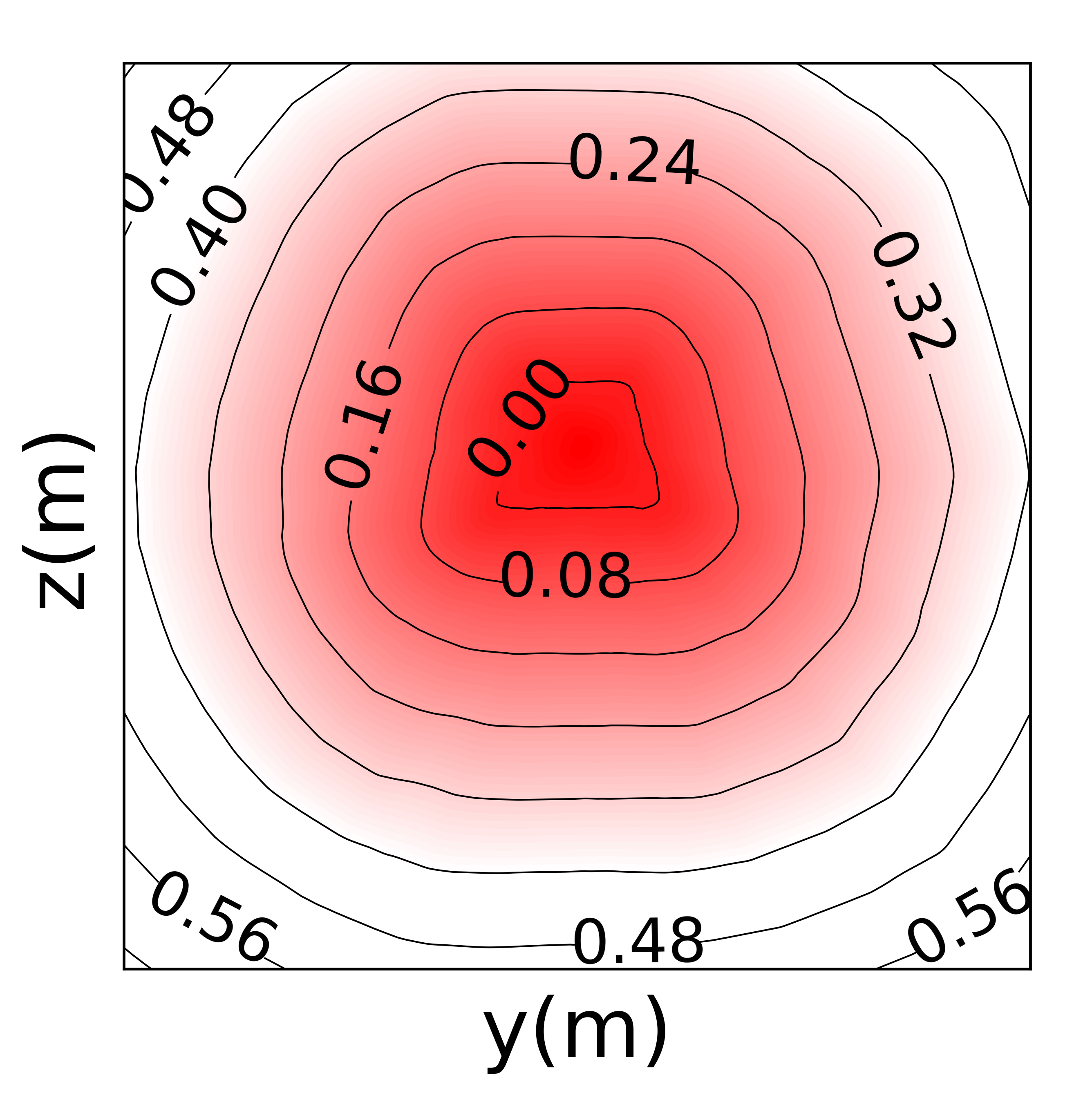}
%\caption{fig2}
\end{minipage}%
}%
\subfigure[]{
\begin{minipage}[t]{0.32\linewidth}
\centering
\includegraphics[width=\textwidth]{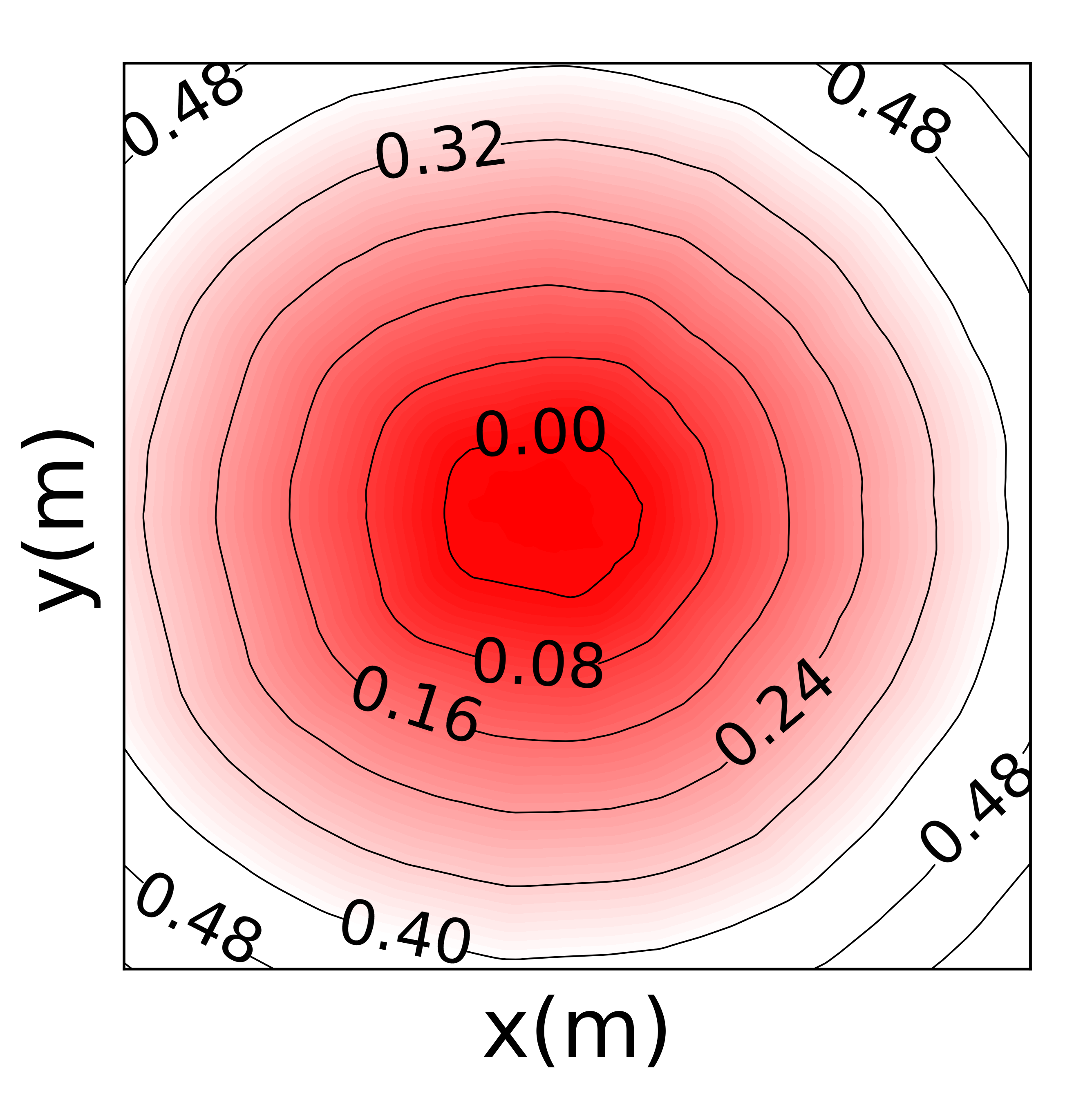}
%\caption{fig2}
\end{minipage}%
}%
\centering
\caption{The reconstructed signed distance field with $f_{\boldsymbol{\theta}}(\mathbf{p})\geq0$ m, for three sections of $link0$ of the Franka Emika Panda robot.}
\label{fig:4}
\end{figure}

\begin{figure}[ht]
\centering
\subfigure[]{
\begin{minipage}[t]{0.3\linewidth}
\centering
\includegraphics[width=\textwidth]{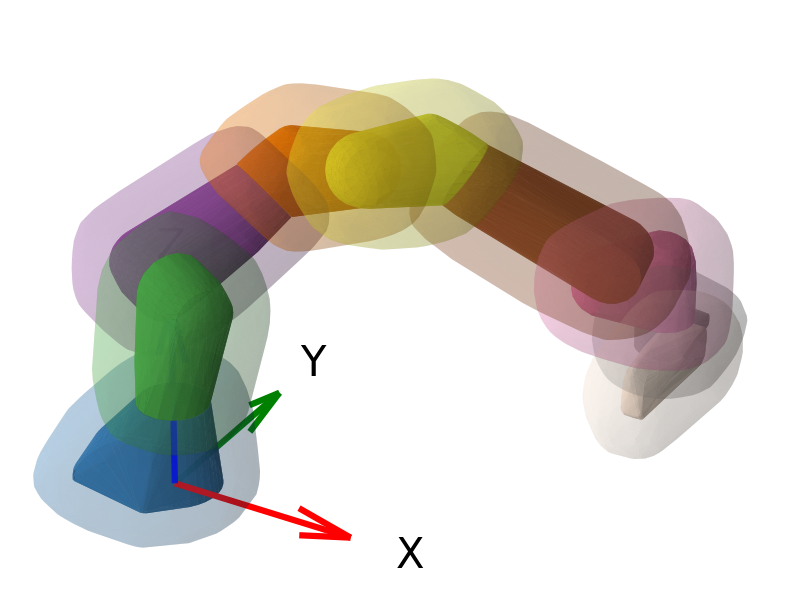}
%\caption{fig1}
\end{minipage}%
}%
\subfigure[]{
\begin{minipage}[t]{0.36\linewidth}
\centering
\includegraphics[width=\textwidth]{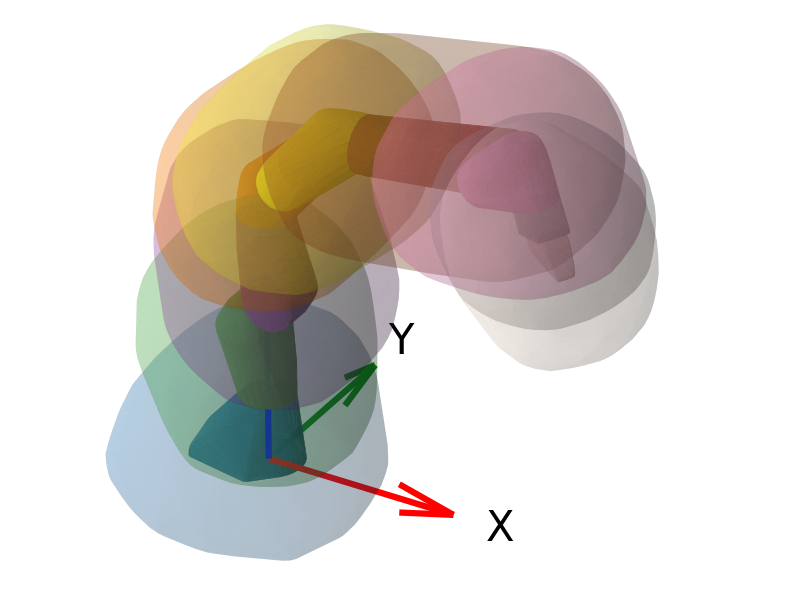}
%\caption{fig2}
\end{minipage}%
}%
\subfigure[]{
\begin{minipage}[t]{0.3\linewidth}
\centering
\includegraphics[width=\textwidth]{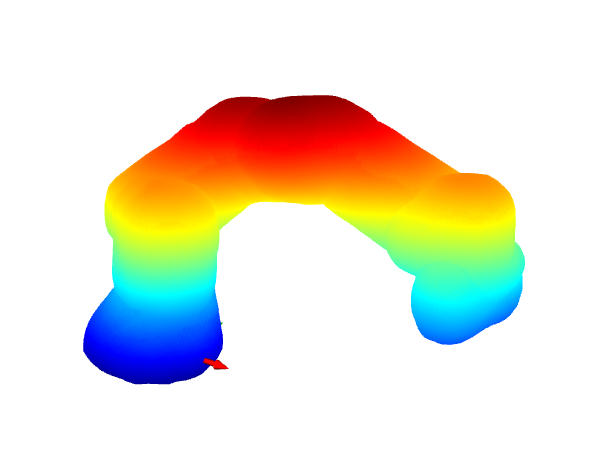}
%\caption{fig2}
\end{minipage}%
}%
\centering
\caption{The learned composite signed distance isosurfaces (transparent) with (a) $f_{\boldsymbol{\theta}}(\mathbf{p}) = 0.05$m and (b) $f_{\boldsymbol{\theta}}(\mathbf{p}) = 0.15$m for each link of the Franka Emika Panda. (c) the merged isosurfaces at $f_{\boldsymbol{\theta}}(\mathbf{p}) = 0.05$m}
\label{fig:5}
\end{figure}

To intuitively demonstrate the accuracy of the learned model, we use the learned SDF network to reconstruct the SDF with $f_{\boldsymbol{\theta}}(\mathbf{p})\geq0$ m, on three sections of the $link0$, i.e., the blue link in Fig. \ref{fig:5}, of the Franka Emika Panda robot, and the results are shown in Fig. \ref{fig:4}. The reconstructed SDF can well reflect the profile of the $link0$ near the surface, and the shape of the isoline tends to be circular as it is far away from the link. Besides, the widths between adjacent isolines are almost equal, also indicating that the learned model can reconstruct an accurate SDF for the robot link.

\begin{table}[ht]
\centering
\caption{Accuracy of the learned SDF network.}
\begin{tabularx}{\linewidth}{l|X|X|X}
\toprule
distance interval (cm) & [0, 40] & [40,80] & [80, 120]  \\
\midrule
RMSD, $ d $ (cm)& 0.28 & 0.36 & 0.38  \\
\midrule
RMSD, $\hat{\mathbf{n}}$ & 0.083 & 0.045 & 0.042 \\
\bottomrule
\end{tabularx}
\label{tab:2}
\end{table}%

% Since we are more interested in the distance accuracy near the robot. Therefore, we evaluate the accuracy close to the robot.

The original mesh geometry (solid) and reconstructed distance isosurface $f_{\boldsymbol{\theta}}(\mathbf{p}) = 0.05$m and $f_{\boldsymbol{\theta}}(\mathbf{p}) = 0.15$m (transparent) of Franka Emika Panda robot at different poses using ray marching algorithms \cite{hart1996sphere}, are shown in Fig. \ref{fig:5}. % the average distance error is $1.32$ mm at this isosurface. 
The reconstructed distance isosurface can well reflect the shape of each link and is not affected by the pose of the robot manipulator. The root-mean-square deviation (RMSD) of the estimated distance $f_{\boldsymbol{\theta}}(\mathbf{p})$ and normal vector $\mathbf{n}$, i.e., the root mean square of the second and third terms in (\ref{eq:5}), of all links at different distance intervals are shown in Table. \ref{tab:2}. Moreover, our method has higher accuracy at two distance intervals, i.e., [0, 10] and [10, 120] cm, and a lower relative network complexity (both are the MLP networks) for the whole robot compared with the most related Neural-JSDF algorithm \cite{koptev2022neural}, and the results are shown in Table. \ref{tab:3}.

\begin{table}[ht]
\centering
\caption{Accuracy Comparison with $0<d<10$ cm $|$ $10<d<120$ cm.}
\begin{tabularx}{\linewidth}{l|X|X}
\toprule
method & composite SDF & Neural-JSDF  \\
\midrule
RMSD, $d$ (cm) & \textbf{0.21} $|$ \textbf{0.36} & 1.04 $|$ 1.06  \\
\midrule
NN complexity, (\%) & 1 & 1.78  \\
\bottomrule
\end{tabularx}
\label{tab:3}
\end{table}%

\subsection{Trajectory Planning Simulation for 2D Mass Point}
\begin{figure}[ht]
\centering
\subfigure[]{
\begin{minipage}[t]{0.32\linewidth}
\centering
\includegraphics[width=\textwidth]{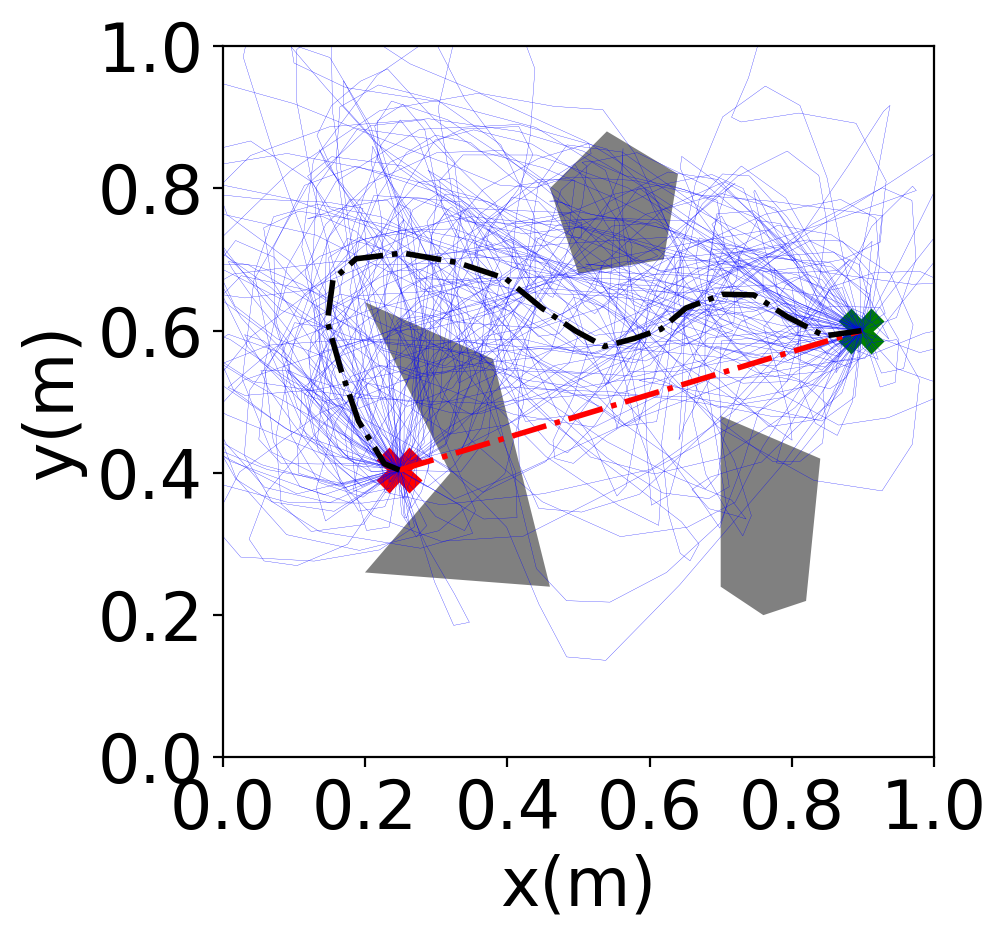}
%\caption{fig1}
\end{minipage}%
}%
\subfigure[]{
\begin{minipage}[t]{0.32\linewidth}
\centering
\includegraphics[width=\textwidth]{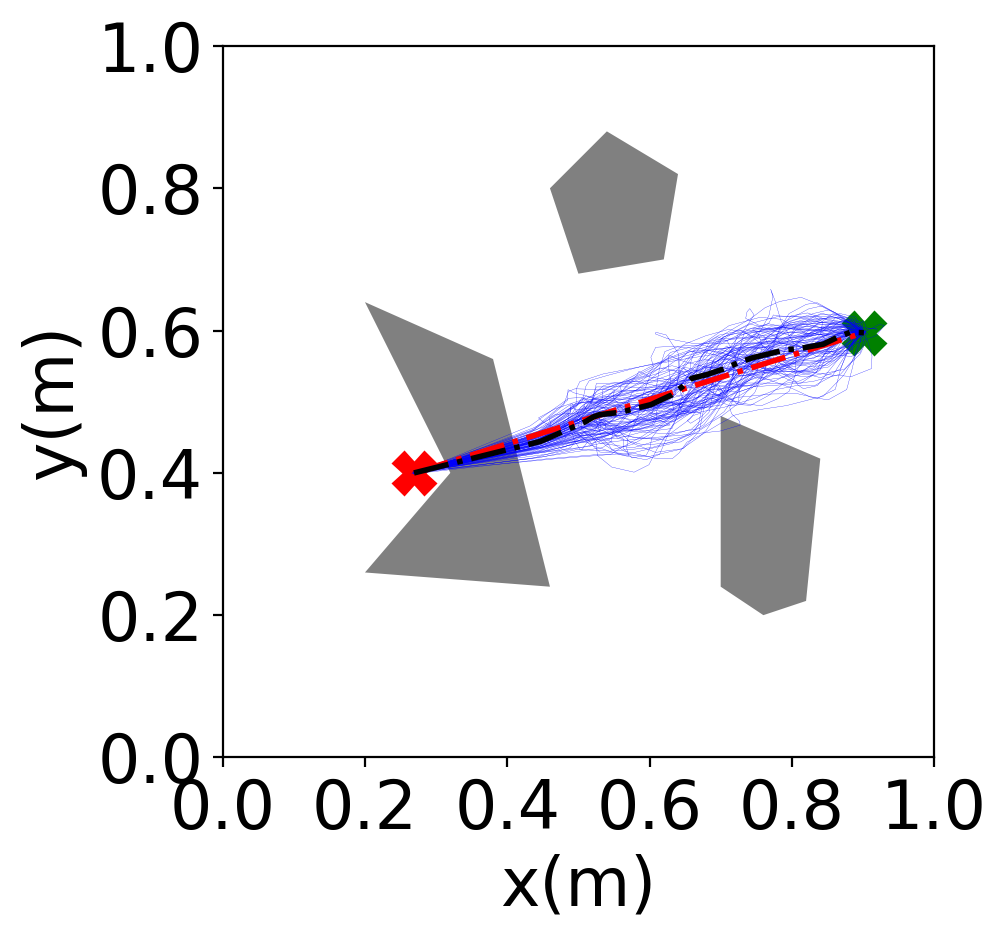}
%\caption{fig2}
\end{minipage}%
}%
\subfigure[]{
\begin{minipage}[t]{0.32\linewidth}
\centering
\includegraphics[width=\textwidth]{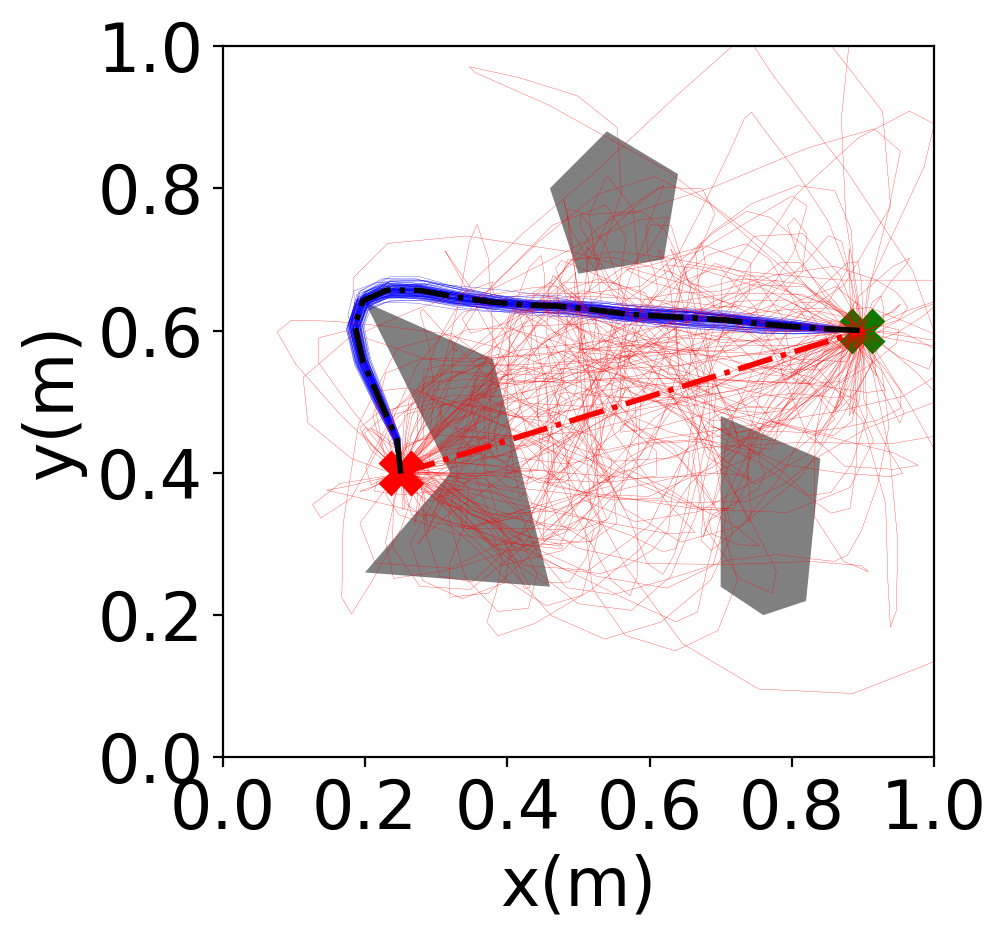}
%\caption{fig2}
\end{minipage}%
}%

\subfigure[]{
\begin{minipage}[t]{0.32\linewidth}
\centering
\includegraphics[width=\textwidth]{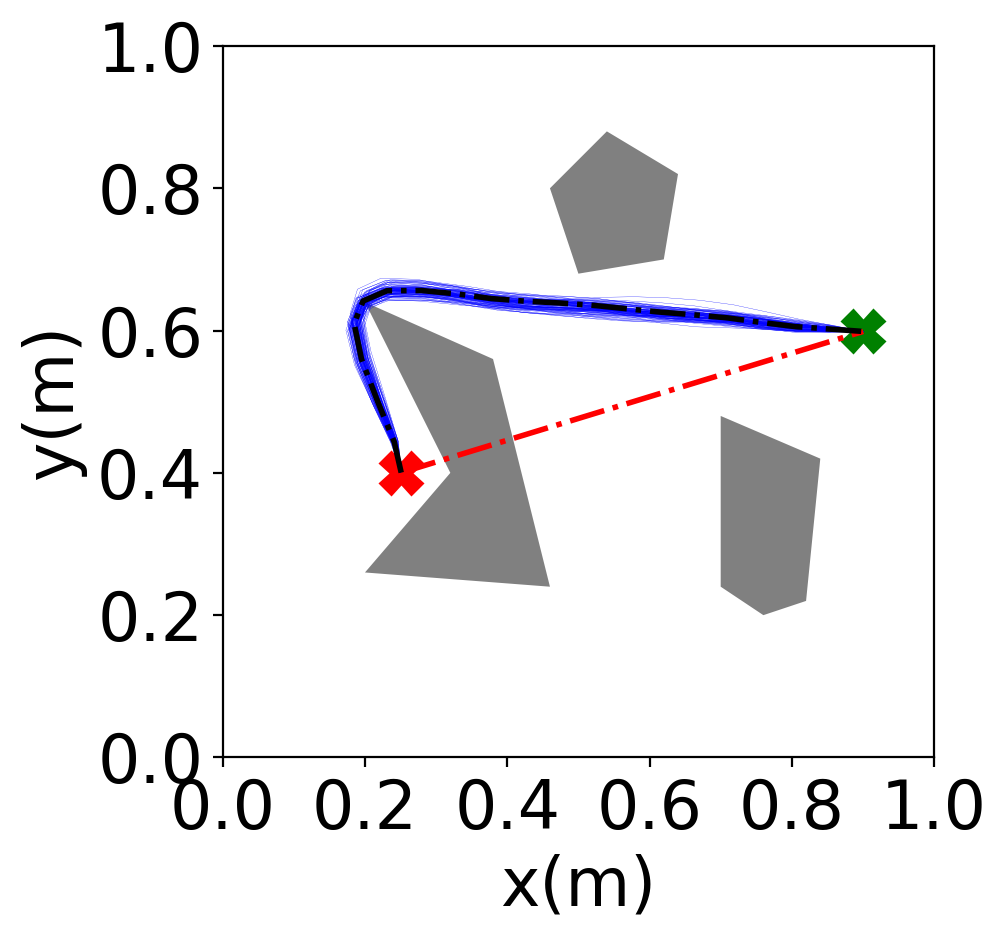}
%\caption{fig2}
\end{minipage}%
}%
\subfigure[]{
\begin{minipage}[t]{0.32\linewidth}
\centering
\includegraphics[width=\textwidth]{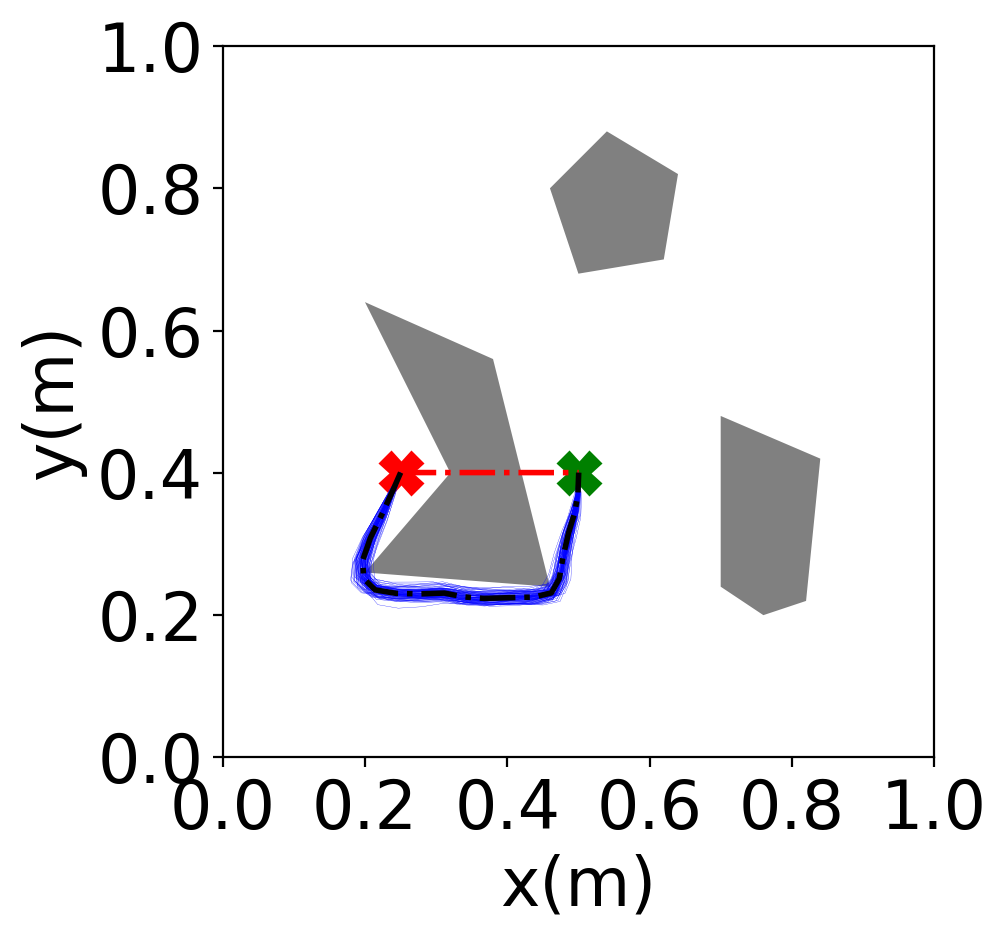}
%\caption{fig1}
\end{minipage}%
}%
\subfigure[]{
\begin{minipage}[t]{0.32\linewidth}
\centering
\includegraphics[width=\textwidth]{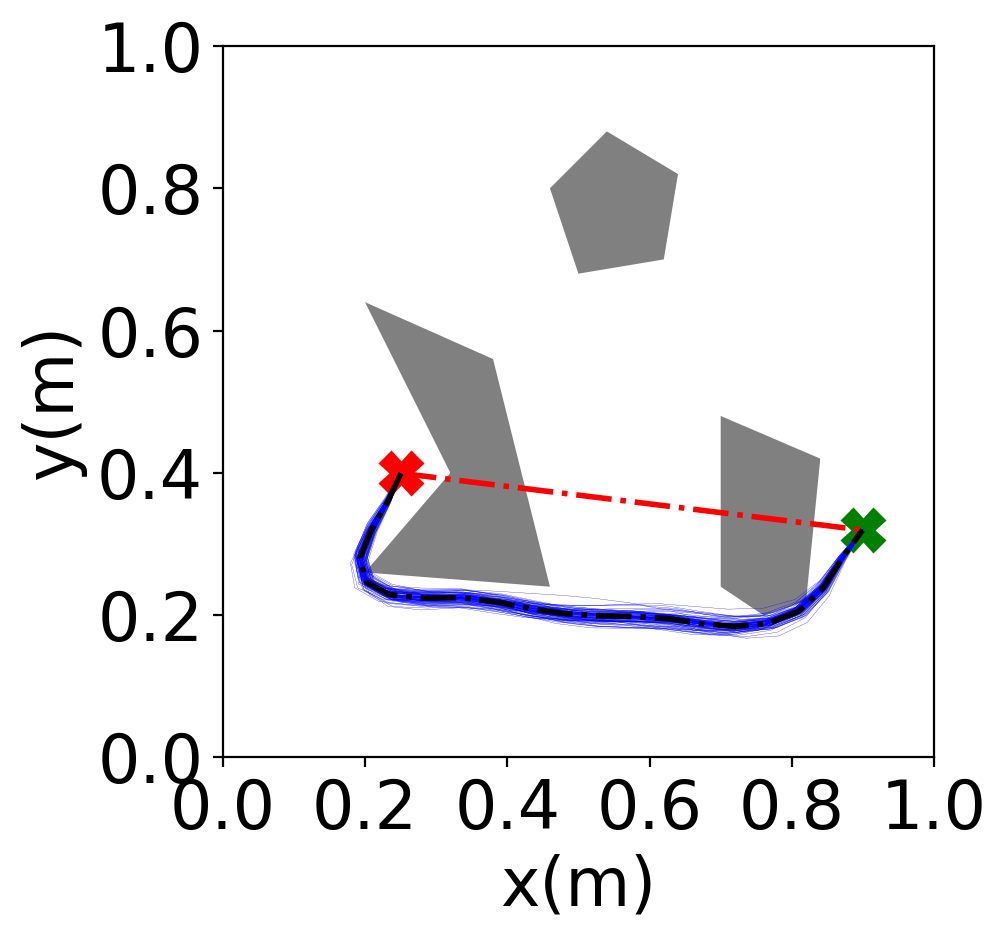}
%\caption{fig2}
\end{minipage}%
}%
\centering
\caption{(a) and (b) denote the generated trajectories of a point mass without using the kernel updating strategies. (c-f) are generated trajectories of a point mass using kernel updating strategies.}
\label{fig:6}
\end{figure}

To verify the effectiveness of the trajectory planning algorithm more intuitively. We first evaluate the algorithm for generating collision-free trajectories for a 2D mass point in a dummy environment as in \cite{bhardwaj2022storm}. In this simulation, the likelihood function consists of collision avoidance and trajectory length penalties. The initial mean trajectory of the Gaussian process prior is a straight line between the start and goal positions as shown by the red dotted lines in Fig. \ref{fig:6}. The discrete steps $H$ of the trajectory are $20$ and the optimization iteration steps are $200$. The parameter $h$ in (\ref{eq:10}) is $0.01$ and $N_s$ in (\ref{eq:17}) is $200$ for all 2D simulation tests.

The black dotted lines in Fig. \ref{fig:6} (a) and (b) are the optimized trajectories without using the kernel updating strategies. The parameter $\sigma_f$ of (a) and (b) are 0.02 and 0.001, respectively. The planning algorithm can generate a collision-free trajectory with $\sigma_f=0.02$ as shown in Fig. \ref{fig:6} (a), but the results are relatively conservative, i.e., the trajectory is far away from obstacles and not short. The planning algorithm cannot generate a collision-free trajectory with $\sigma_f=0.001$ as shown in Fig. \ref{fig:6} (b), and sticks into a local optimum. This comparison shows that larger $\sigma_f$ is beneficial to explore collision-free trajectories, and the trajectory is easy to stick into the local optimum with smaller $\sigma_f$. For the 2D point mass simulation demonstrations, one can refer to the video in the attachment.

The initial and optimized trajectories of point mass using the kernel updating strategies are shown in Fig. \ref{fig:6} (c-f). All of them are set with a large initial $\sigma_{f}=0.02$ and minimum allowable $\sigma_{min}=0.0005$ and the corresponding trajectory samples are shown by the red and blue solid lines in Fig. \ref{fig:6} (c), respectively. The difference between (c) and (d) is that $5$ linear interpolation points between adjacent waypoints are taken into consideration in (d) while estimating the cost for obstacle avoidance. All the $20$ waypoints in (c) and (d) are located in the collision-free white area.  As the goal position changes, the trajectories to be optimized have different lengths and complexities. In both cases, the proposed trajectory planning algorithm can generate collision-free and short trajectories regardless of the trajectory length, as shown in Fig. \ref{fig:6} (e) and (f).

% \begin{table}[ht]
% \centering
% \caption{Comparison with other trajectory planning algorithms.}
% \begin{tabularx}{\linewidth}{l|X|X|X|X}
% \toprule
% method & STP-KUS & TrajOpt & GPMP \\
% \midrule
% success rate, (\%) & 20 & 40 & 20  \\
% \bottomrule
% \end{tabularx}
% \label{tab:4}
% \end{table}%
Besides, since the initial and target positions of the mass point are placed on both sides of the concave obstacle, as shown in Fig. \ref{fig:6} (e). In such a dilemma, using the gradient descent-based trajectory planning algorithms in \cite{lambert2021entropy} and \cite{schulman2013finding} with an initial guess of a straight line will only converge to this invalid local optimum. The cost for collision avoidance, i.e., intuitively, it can be regarded as the overlapping length of the trajectory and the obstacle, will increase in both directions, i.e., up and down. The results demonstrate that the proposed trajectory planning algorithm has a certain ability to escape from the local optimum.

\subsection{Evaluation of the Whole Algorithms}
In this part, the learned composite SDF networks are integrated into the trajectory planning algorithm, and the whole algorithm is tested on the Franka Emika Panda robot manipulator to verify the effectiveness and feasibility of the algorithm.

\subsubsection{Testing Scenarios and Parameter Settings}
The robot manipulator attempts to perform tasks of goal-reaching in Cartesian space while avoiding collision with obstacles. Since the Gaussian process prior is constructed in the configuration space, Inverse Kinematics (IK) is used to estimate the target configuration. The workbench consists of a 7-DOF Franka Emika Panda robot manipulator with a fixed base and a Kinect-V2 camera mounted on a fixed position to detect obstacles. In this case, the Aruco markers are used for hand-eye calibration and obstacle position detection.

The entire trajectory is discretized into $20$ waypoints, i.e., $H=20$ steps, and as calculating the cost for obstacle avoidance, two interpolation points between adjacent waypoints are considered. In addition to the cost for obstacle avoidance and trajectory length penalty, the likelihood function also contains a cost for boundary avoidance that constrains the $z$-position of the robot, i.e., the $z$-position of the link should not be less than $0.02$ m. The distance threshold is $\epsilon = 0.08$ m. The number of particles used to update the mean trajectory is $N_s=50$. The initial value of $\sigma_{f}=0.1$ and the minimum allowable $\sigma_{f}=0.012$. The length scale is $l=0.5$. The joint speed and acceleration limits of the robot for the TOPP-RA-based time-optimal planner are set to $0.1$ times the preset limit value of the Franka Emika Panda robot.
\begin{figure}[ht]
\centering
\subfigure[]{
\begin{minipage}[t]{0.32\linewidth}
\centering
\includegraphics[scale=0.085]{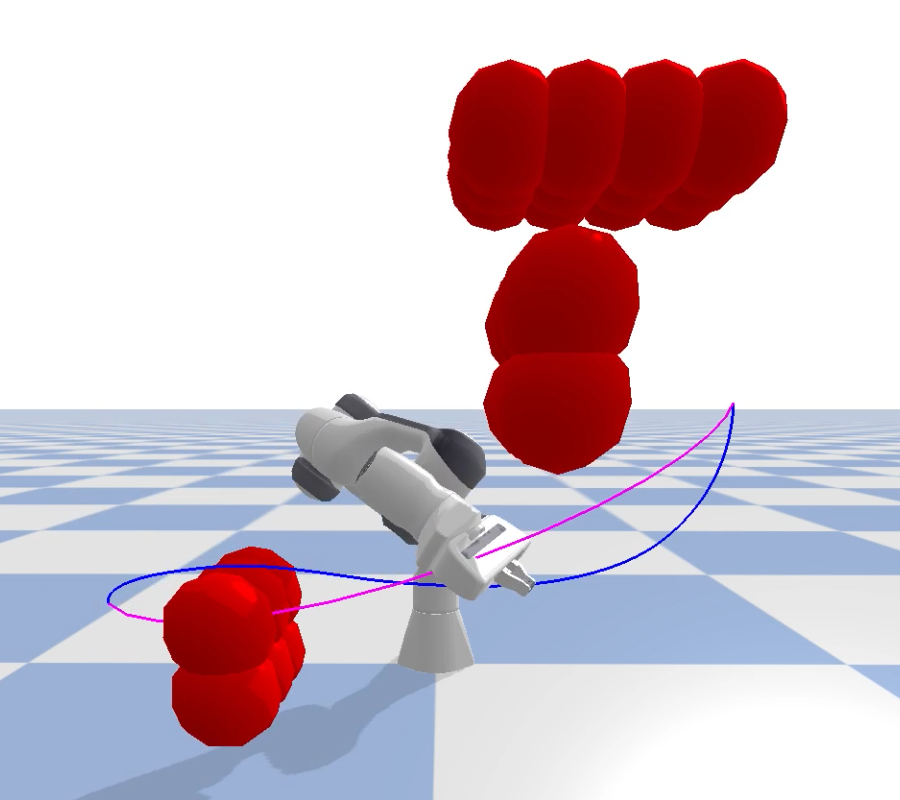}
%\caption{fig1}
\end{minipage}%
}%
\subfigure[]{
\begin{minipage}[t]{0.32\linewidth}
\centering
\includegraphics[scale=0.085]{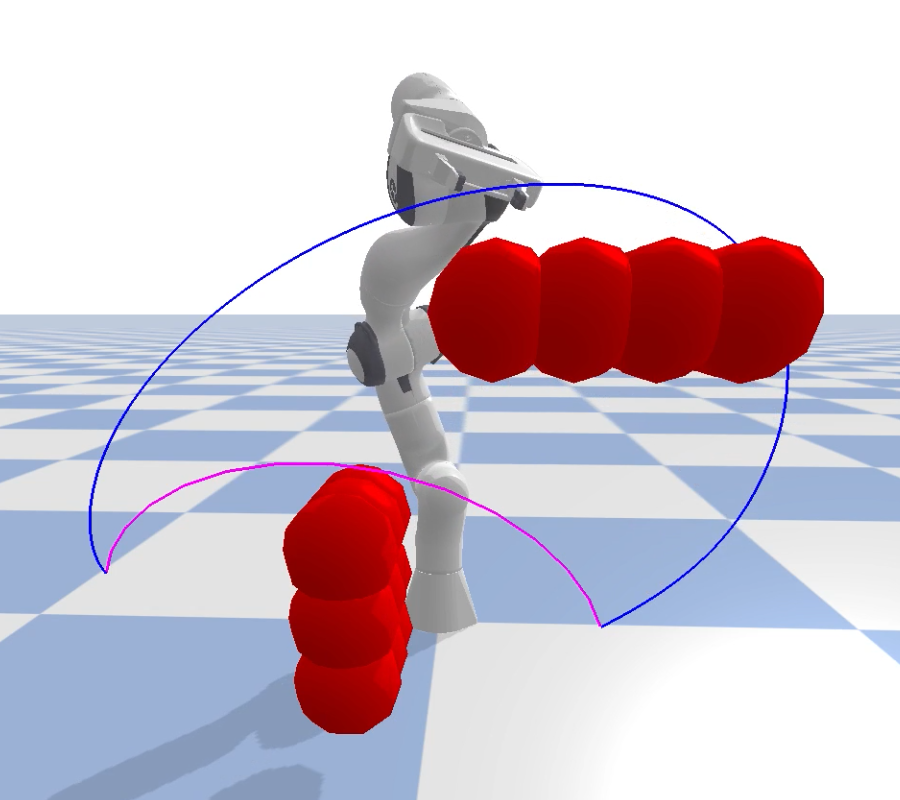}
%\caption{fig2}
\end{minipage}%
}%
\subfigure[]{
\begin{minipage}[t]{0.32\linewidth}
\centering
\includegraphics[scale=0.085]{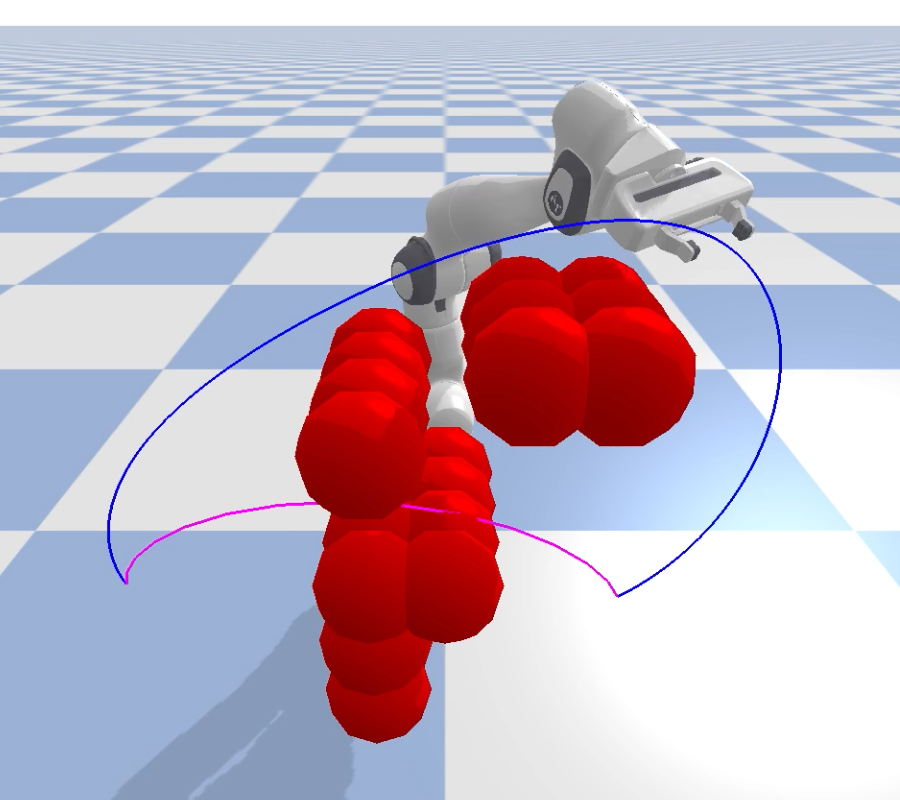}
%\caption{fig2}
\end{minipage}%
}%
\centering
\caption{Screenshot of three different obstacle avoidance scenarios. The pink curve is the initial mean trajectory, and the blue curve is the optimized collision-free trajectory.}
\label{fig:7}
\end{figure}

\begin{figure}[ht]
\centering
\subfigure[]{
\begin{minipage}[t]{0.23\linewidth}
\centering
\includegraphics[scale=0.065]{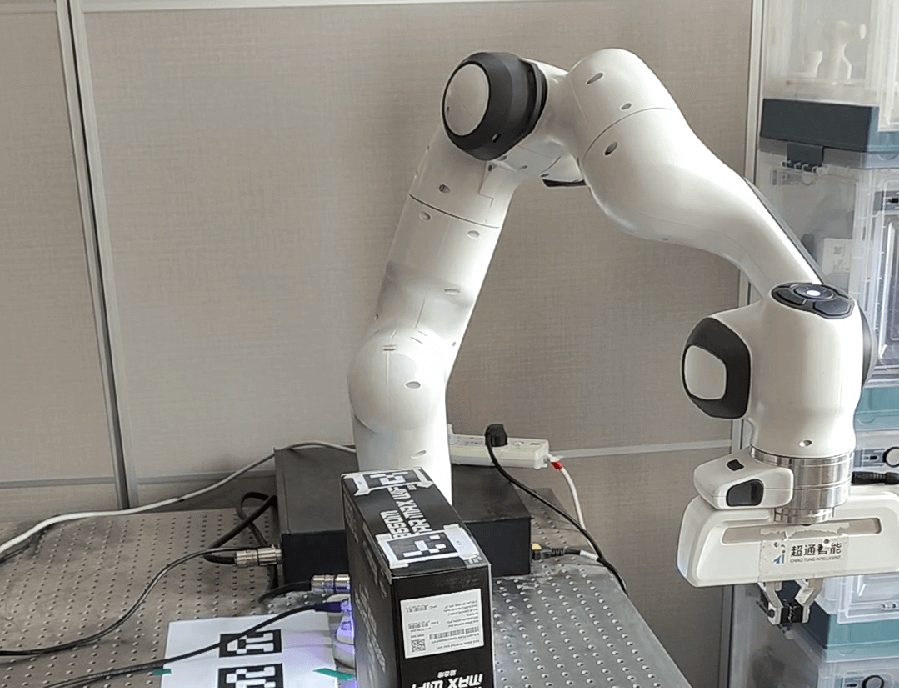}
%\caption{fig1}
\end{minipage}%
}%
\subfigure[]{
\begin{minipage}[t]{0.23\linewidth}
\centering
\includegraphics[scale=0.065]{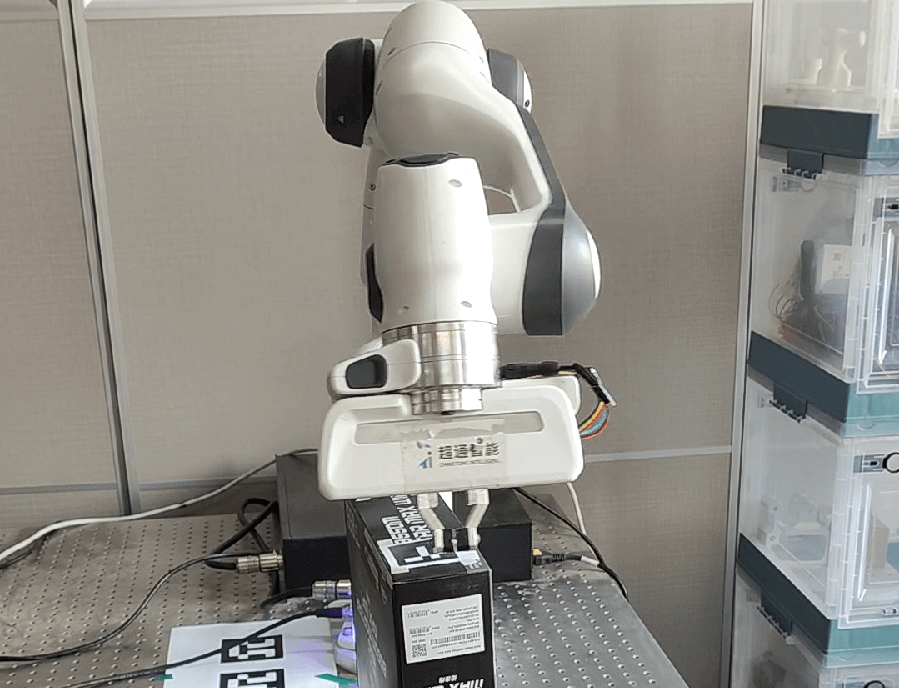}
%\caption{fig2}
\end{minipage}%
}%
\subfigure[]{
\begin{minipage}[t]{0.23\linewidth}
\centering
\includegraphics[scale=0.065]{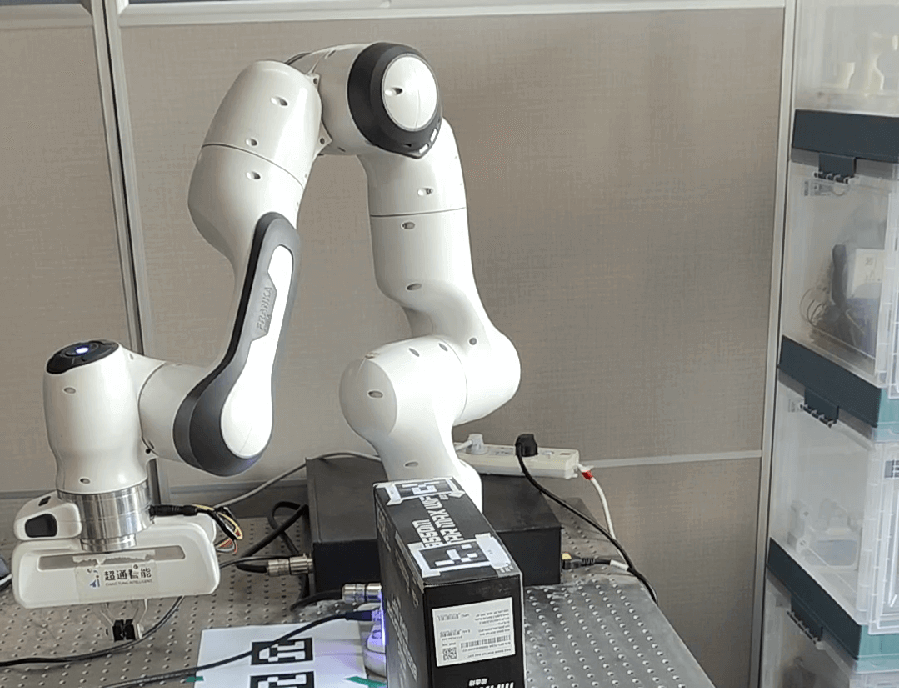}
%\caption{fig2}
\end{minipage}%
}%
\subfigure[]{
\begin{minipage}[t]{0.23\linewidth}
\centering
\includegraphics[scale=0.117]{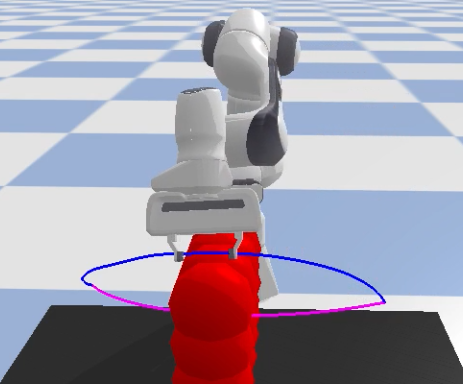}
%\caption{fig2}
\end{minipage}%
}%
\centering
\caption{Screenshot of Franka Emika Panda robot performing collision avoidance motion.}
\label{fig:8}
\end{figure}

\subsubsection{Evaluation Results}
First, the proposed algorithms are applied to plan collision-free trajectories in the three simulation scenarios as shown in Fig. \ref{fig:7}. The pink curves denote the initial mean trajectory of the end-effector, and the blue curve is the trajectory obtained after optimization. The robot generates collision-free trajectories in all three scenarios. For the demonstrations, please refer to the attached video.

\begin{figure}[ht]
\centering
\includegraphics[scale=0.75]{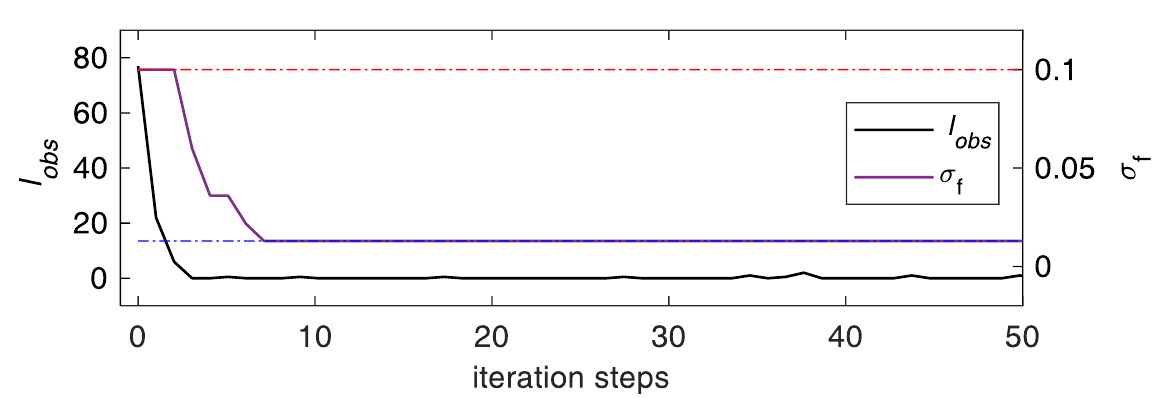}
\caption{The curve of cost for obstacle avoidance $\ell_{obs}$ corresponding to the mean trajectory and the evolution curve of $\sigma_{f}$ during the iteration process. }
\label{fig:9}
\end{figure}

Then, the proposed algorithms are tested in a scenario robot, as shown in Fig. \ref{fig:8} (a) to (c). In the robot workspace, the black box is an obstacle that is simplified as a set of spheres. The manipulator needs to move from the right side of the box to its left side. The initial (pink curve) and the optimized (blue curve) trajectory of the end-effector are shown in  Fig. \ref{fig:8} (d). The robot can move from the initial configuration to the goal pose while avoiding collision with the black box, as shown in Fig. \ref{fig:8}. Fig. \ref{fig:9} denotes the curve of the cost for obstacle avoidance corresponding to the mean trajectory and the evolution curve of $\sigma_{f}$ during the iteration process. The cost for obstacle avoidance quickly converges to 0, i.e., after 3 iterations as shown by the black solid line in Fig. \ref{fig:9}. $\sigma_{f}$ decreases from 0.1 to 0.012 after the cost for obstacle avoidance reaches zero as shown by the purple solid line in Fig. \ref{fig:9}. The robot can avoid collision with the black box and generate collision-free trajectories from the results which also shows that the whole algorithm is effective. For the robot demonstrations, please refer to the attached video.

\section{Conclusion}

We have proposed composite SDF networks for articulated robots, which can  quickly and accurately calculate the minimum distance between the articulated robot and obstacles combined with the kinematics of the articulated robot. Although the learned composite SDF networks contain many parameters, it still runs fast due to the high degree of parallelization. In addition, we also propose a stochastic trajectory optimization method based on the Gaussian process prior and kernel function update strategy, which can efficiently generate goal-directed motion trajectories for high-dimensional robots. The introduction of the MPPI with kernel update strategy to explore collision-free trajectories endows the optimization process with a certain ability to escape from local optimum and does not generate overly conservative trajectories. In addition, the composite SDF model is integrated into the trajectory generation algorithm which can generate optimized and collision-free trajectories for high degree freedom robot manipulator.

Although the composite SDF networks are used for offline trajectory planning, this network still runs fast for real-time distance calculation and can be used for obstacle avoidance algorithms in dynamic environments. Thus, in the future, we would also like to integrate the composite SDF networks into a real-time motion planning and control method to ensure collision-free motion generation in a dynamic environment. The limitations of this network include that it does not take into account relative velocities between robots and obstacles, nor does it include information on gradients of distances with respect to robot joint positions.

% or


\ifCLASSOPTIONcaptionsoff
  \newpage
\fi



%\bibliographystyle{IEEEtran}
\bibliographystyle{IEEEtran}
\bibliography{ref/safety_motion_planning}\



%\newpage


%\vfill


\end{document}